%% file: main.tex
\documentclass{sigchi-ext}

\usepackage[T1]{fontenc}
\usepackage{textcomp}
\usepackage[scaled=.92]{helvet}
\usepackage{graphicx}
\usepackage{balance}
\usepackage{booktabs}
\usepackage{adjustbox}
\usepackage{ccicons}
\usepackage{ragged2e}
\usepackage{caption}
\usepackage{tikz}

\title{Massif: Interactive Interpretation of Adversarial Attacks on Deep Learning}
\input{001-authors}

\input{002-defs}
\input{003-copyright}

\begin{document}

\maketitle
\RaggedRight{}

\input{010-abstract}
\input{011-keywords}

\input{100-intro}
\input{300-system}
\input{400-result}
\input{500-conclusion}

\section{Acknowledgements}
This work was supported in part by NSF grants IIS-1563816, CNS-1704701, NASA NSTRF, DARPA GARD, gifts from Intel (ISTC-ARSA), NVIDIA, Google, Symantec, Yahoo! Labs, eBay, Amazon.

\balance{} 

\bibliographystyle{SIGCHI-Reference-Format}
\bibliography{ref}

\end{document}

%% file: 001-authors.tex
\numberofauthors{9}

\author{
  \alignauthor{
    \textbf{Nilaksh Das*} \\
    \affaddr{Georgia Institute of Technology} \\
    \affaddr{Atlanta, GA, USA} \\
    \email{nilakshdas@gatech.edu}
  }
  \alignauthor{
    \textbf{Robert Firstman}\\
    \affaddr{Georgia Institute of Technology} \\
    \affaddr{Atlanta, GA, USA}\\
    \email{rfirstman6@gatech.edu} 
  }\vfil 
  \alignauthor{
    \vspace{1em}
    \textbf{Haekyu Park*}\\
    \affaddr{Georgia Institute of Technology}\\
    \affaddr{Atlanta, GA, USA}\\
    \email{haekyu@gatech.edu} 
  }
  \alignauthor{
    \vspace{1em}
    \textbf{Emily Rogers}\\
    \affaddr{Georgia Institute of Technology} \\
    \affaddr{Atlanta, GA, USA}\\
    \email{Emily.Rogers@gtri.gatech.edu}
  }\vfil 
  \alignauthor{
    \vspace{1em}
    \textbf{Zijie J. Wang}\\
    \affaddr{Georgia Institute of Technology}\\
    \affaddr{Atlanta, GA, USA}\\
    \email{jayw@gatech.edu} 
  }
  \alignauthor{
    \vspace{1em}
    \textbf{Duen Horng (Polo) Chau}\\
    \affaddr{Georgia Institute of Technology}\\
    \affaddr{Atlanta, GA, USA}\\
    \email{polo@gatech.edu}
  }
  \alignauthor{
    \vspace{1em}
    \textbf{Fred Hohman}\\
    \affaddr{Georgia Institute of Technology}\\
    \affaddr{Atlanta, GA, USA}\\
    \email{fredhohman@gatech.edu} 
  }
  \alignauthor{
    \textbf{$\;$}\\
  }
  \alignauthor{
    \textbf{$\;$}\\
    \affaddr{$\;$}\\
    \affaddr{*Authors contributed equally.}\\
  }
}

%% file: 002-defs.tex
\newcommand{\model}[0]{\textsc{Massif}}

\definecolor{orange}{RGB}{255,119,0}
\definecolor{purple}{RGB}{158, 62, 177}
\definecolor{lightblue}{RGB}{72, 123, 232}

\newcommand{\blue}[1]{\textcolor{lightblue}{#1}}

\newcommand{\orange}[1]{\textcolor{orange}{#1}}
\newcommand{\purple}[1]{\textcolor{purple}{#1}}

%% file: 003-copyright.tex
\copyrightinfo{\scriptsize Permission to make digital or hard copies of part or all of this work for personal or classroom use is granted without fee provided that copies are not made or distributed for profit or commercial advantage and that copies bear this notice and the full citation on the first page. Copyrights for third-party components of this work must be honored. For all other uses, contact the owner/author(s). \\
{\emph{CHI '20 Extended Abstracts, April 25--30, 2020, Honolulu, HI, USA.}} \\
\copyright~2020 Copyright is held by the author/owner(s). \\ 
ACM ISBN 978-1-4503-6819-3/20/04. \\
http://dx.doi.org/10.1145/3334480.3382977}

%% file: 010-abstract.tex
\begin{abstract}

Deep neural networks (DNNs) are increasingly powering high-stakes applications such as autonomous cars and healthcare; however, DNNs are often treated as ``black boxes'' in such applications.
Recent research has also revealed that DNNs are highly vulnerable to adversarial attacks, raising serious concerns over deploying DNNs in the real world.
To overcome these deficiencies, we are developing \model{}, an interactive tool for deciphering adversarial attacks.
\model{} identifies and interactively visualizes neurons and their connections inside a DNN that are strongly activated or suppressed by an adversarial attack.
\model{} provides both a high-level, interpretable overview of the effect of an attack on a DNN, and a low-level, detailed description of the affected neurons. 
\model{}'s tightly coupled views help people better understand which input features are most vulnerable and important for correct predictions.

\end{abstract}

%% file: 011-keywords.tex
\keywords{
    Deep learning interpretability,
    adversarial attack,
    visual analytics, 
    scalable summarization,
    attribution graph
}

%% file: 100-intro.tex
\section{Introduction}

Deep neural networks (DNNs) have demonstrated significant success in a wide spectrum of applications \cite{esteva2019guide,grigorescu2019survey,guo2019survey,nassif2019speech}.
However, they have been found to be highly vulnerable to \textit{adversarial attacks}: typically small, human-imperceptible perturbations on inputs that fool DNNs into making incorrect predictions \cite{chen2018shapeshifter,Goodfellow2014ExplainingAH,kurakin2016adversarial,qin2019imperceptible}.
This jeopardizes many DNN-based technologies, especially in security and safety-critical applications such as autonomous driving and data-driven healthcare.
To make deep learning more robust against such malicious attacks, it is essential to understand how the attacks permeate DNN models \cite{ross2018improving,tao2018attacks}. 
Interpreting, and ultimately defending against adversarial attacks, is nontrivial due to a number of challenges:

\begin{enumerate}[topsep=0mm, itemsep=0mm, parsep=2mm, leftmargin=7mm]

\item[C1.]{\textbf{Scaling to large datasets.}}
Since DNNs contain thousands of neurons and connections, summarizing how an attack affects a model requires analyzing the model's behavior over entire datasets and developing scalable representations that characterize the attack~\cite{carter2019activation, hohman2019summit}.

\item[C2.]{\textbf{Entangled features and connections between benign and attacked inputs.}} 
A natural approach for understanding adversarial attacks is to compare a model's operations on benign and attacked inputs, which could help people understand \textit{where} and \textit{why} predictions within a model diverge.
However, designing an effective comparison can be challenging because the features contributing to the differences may correlate with one or more features in both the benign and attacked classes.
For example, \autoref{fig:intro-compare} shows that a feature representing ``ivory face with dark eyes and nose'' (center purple node) is important for both the benign \textit{panda} class and the attacked \textit{armadillo} class.
This shared feature correlates with a feature for panda (e.g., ``black \& white patches''), 
while also correlating with a few features for armadillo (e.g., ``scales'', ``crossed pattern'').

\begin{figure}[t]
    \centering
    \includegraphics[width=\linewidth]{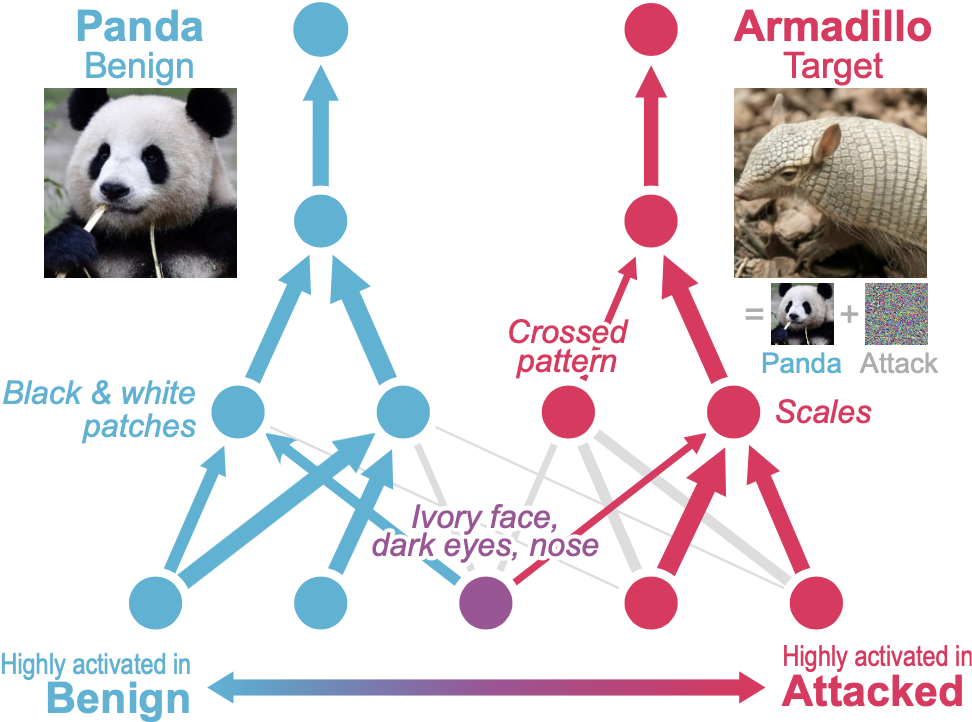}
    \caption{
        Adversarial attacks confuse DNNs to make incorrect predictions, e.g., attacking benign \textit{panda} images so they are misclassified as \textit{armadillo}.
        We aim to understand where such attacks occur inside the model and what features are used.
    }
    \label{fig:intro-compare}
    \vspace{-1em}
\end{figure}

\item[C3.]{\textbf{Diverse feature vulnerability.}}
Given an adversarial attack, some learned features may be more easily manipulated and vulnerable than others.
For example, manipulating a ``basketball'' feature into an ``orange'' feature is easier (similar shape and color) than changing it into a ``truck'' feature.  
As features exhibit a spectrum of vulnerability, enabling users to visualize and understand an attack under varying levels of severity could help them design stronger countermeasures.
\end{enumerate}

\subsection{Contributions}
To address the aforementioned challenges, we are developing \textbf{\model{}}, an interactive visualization tool for interpreting adversarial attacks on deep learning models.
Our ongoing work presents the following contributions: 

\begin{enumerate}[topsep=0mm, itemsep=0mm, parsep=2mm, leftmargin=5mm]

\item \textbf{Novel graph-based comparison.} 
To discover the features and connections activated or suppressed by an attack, we adapt the recently proposed \textit{attribution graph}~\cite{hohman2019summit} in a \textbf{novel way to visualize, summarize, and compare a model's response to benign and attacked data}.
The original attribution graph aims to highlight how a model's learned features interact to make predictions for a single class, by representing highly activated neurons as vertices and their most influential connections as edges.
Our main idea for \model{} is to generate and integrate two attribution graphs: one for the benign data and another for the attacked data, as illustrated in \autoref{fig:intro-compare}.
The aggregated graph helps us understand which features are shared by both benign and attacked data (e.g., purple, center feature in \autoref{fig:intro-compare}), which are solely activated by the benign data (blue, far left), and which are by the attacked data (red, far right).
Importantly, \model{} also helps users more easily discover where a prediction starts to ``diverge'', honing in to the \textbf{critical parts} of the model that the attack is exploiting.

\item{\textbf{Fractionation of neurons based on vulnerability.}}
To help users prioritize their inspection of neurons, we develop a new way to sort and group them based on their vulnerability, i.e., ``how easily can a neuron be activated or suppressed by an attack.''
Our main idea is to vary an attack's strength (or severity) and record all neuron activations.
Neurons that are easily activated or suppressed by solely a weak attack may warrant focused inspection since they can be easily manipulated with little effort. 
\end{enumerate}

\begin{marginfigure}[-30pc]
  \begin{minipage}{\marginparwidth}
    \centering
    \includegraphics[width=1.0\marginparwidth]{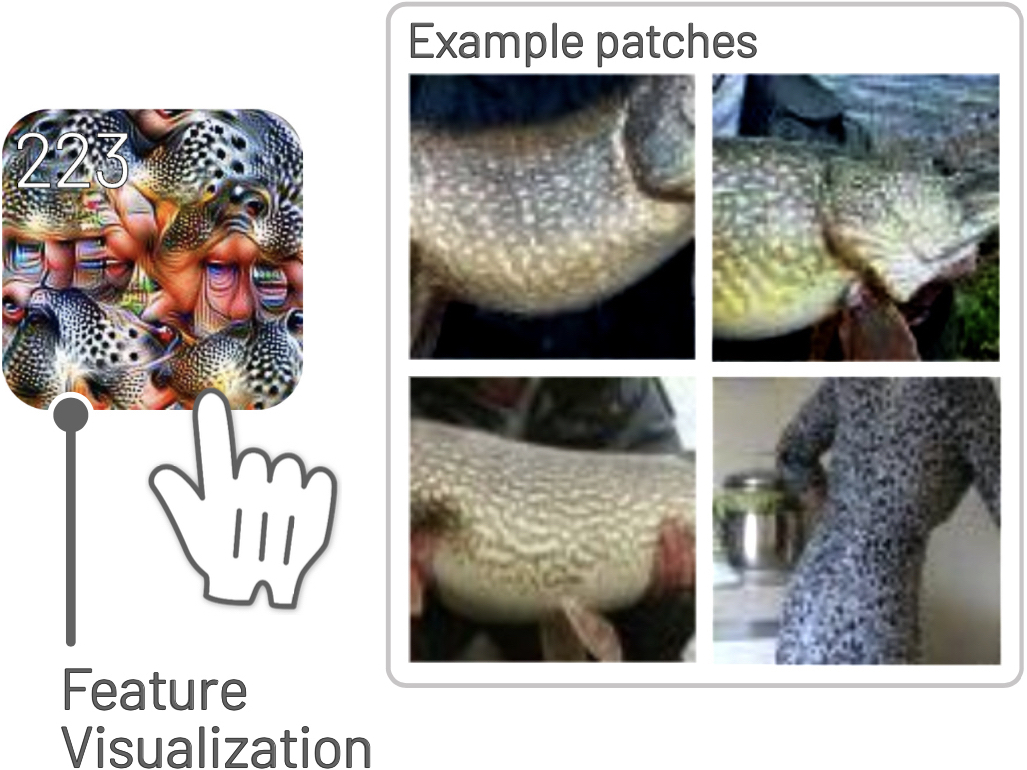}
    \caption{
      Each neuron in an attribution graph is represented with its feature visualization.
      When a user hovers over a neuron, example dataset patches are shown for context. 
    }
    \label{fig:node}
  \end{minipage}
\end{marginfigure}

\begin{figure*}
\begin{minipage}[t]{0.925\marginparwidth}
  \hspace*{-13em}
  \parbox{0.925\marginparwidth}{
  }
\end{minipage}%
\begin{minipage}[t]{\textwidth}
  \centering
  \hspace*{-1.85\marginparwidth}
  \parbox{\textwidth}{
    \includegraphics[width=\textwidth]{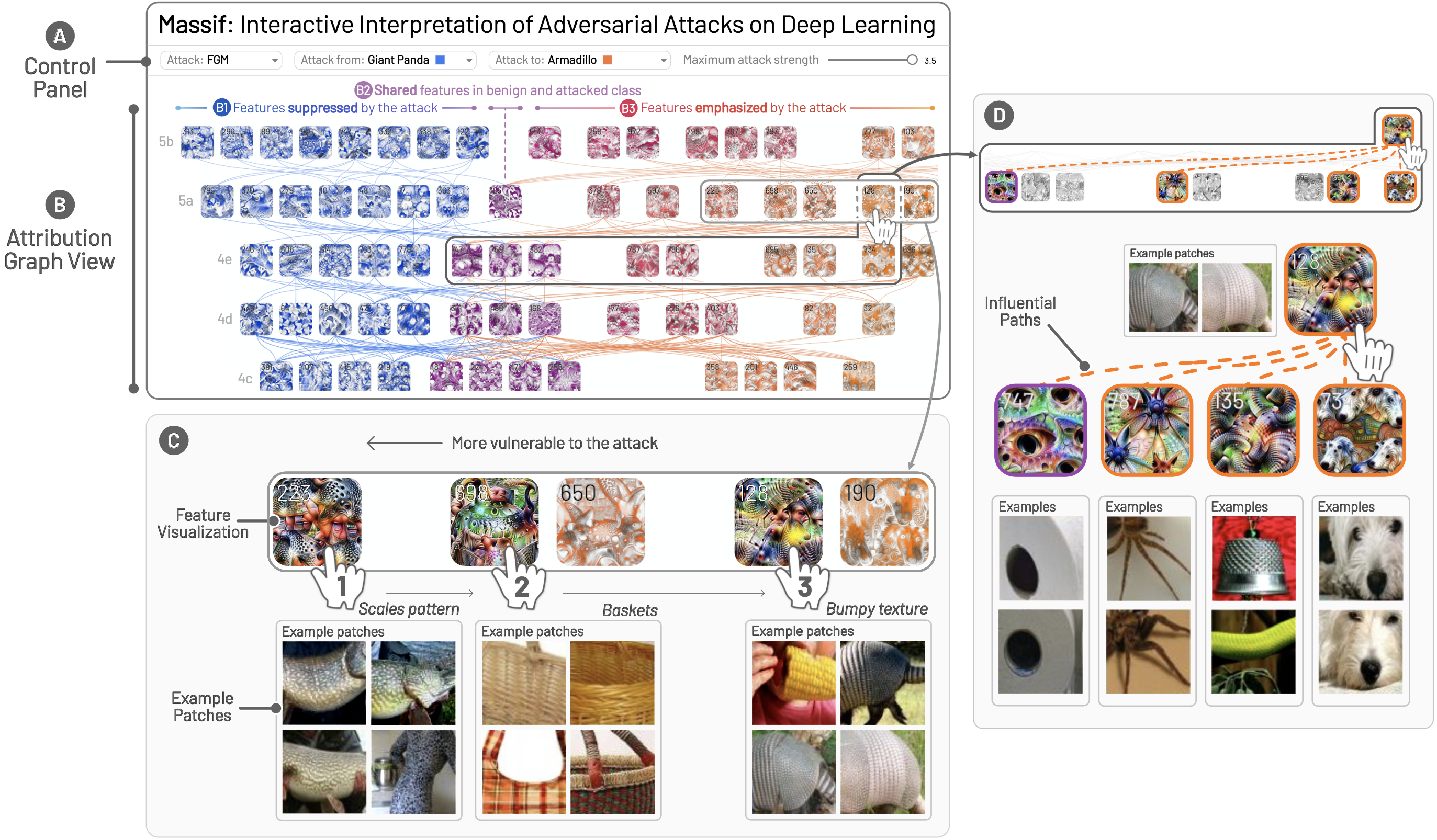}
    \captionsetup{labelformat=empty}
    \parbox{\textwidth}{
      \vspace{1em}
      \textbf{Figure~\ref{fig:overview}:}
      The \model{} interface.
      A user Hailey is studying the targeted Fast Gradient Method (FGM) attack performed on the InceptionV1 model. 
      Using the control panel \textbf{(A)}, she selects ``giant panda'' as the benign class and ``armadillo'' as the attack target class.
      \model{} generates an attribution graph \textbf{(B)}, which shows Hailey the neurons within the network that are suppressed in the attacked images (B1, \blue{\textbf{blue}}), shared by both benign and attacked images (B2, \textbf{\purple{purple}}), and emphasized only in the attacked images (B3, \textbf{\orange{orange}}). 
      Each neuron is represented by a node and its feature visualization \textbf{(C)}.
      Hovering over any neuron displays example dataset patches that maximally activate the neuron, providing stronger evidence for what a neuron has learned to detect. 
      Hovering over a neuron also highlights its most influential connections from the previous layer \textbf{(D)}, allowing Hailey to determine where in the network the prediction diverges from the benign class to the attacked class.
    }
    \caption{}\label{fig:overview}
  }
\end{minipage}
\end{figure*}

%% file: 300-system.tex
\section{System Design and Implementation}
This section describes \model{}'s interface and visualization design (see \autoref{fig:overview}).
Users can select a benign class, an attacked class, and the type and severity of an attack from the top control panel (\autoref{fig:overview}A).
Then, \model{} generates attribution graphs using a Python backend, and displays them to the user in a web-based interface built with HTML, CSS, JavaScript, and D3 (\autoref{fig:overview}B).

\paragraph{Nodes: Features activated (or suppressed) by attacks.}
In attribution graphs, nodes represent DNN neurons which are trained to detect particular features in input data.
To interpret what features a neuron detects, \model{} represents each neuron with its feature visualization: a synthesized image that maximizes the neuron's activation~\cite{olah2017feature}.
Users can hover over any neuron's feature visualization to also display example image patches from the dataset that most activate that neuron.
For example, as seen in \autoref{fig:node}, the feature visualization (left) and dataset examples (right) describe a neuron that detects a dotted pattern in scales.

We divide an attribution graph's neurons into three groups by their attack response.
First, \textbf{suppressed neurons} are highly activated by benign inputs but become suppressed by adversarial inputs.
These represent crucial features for the benign class, but the model fails to detect them when exposed to the attack.
Second, \textbf{emphasized neurons} are not noticeably activated by benign inputs but become highly activated by adversarial inputs.
These represent features that are typically not important for the benign class, but the model detects them as important features of the attacked class.
Third, \textbf{shared neurons} are highly activated by and important to both benign and adversarial inputs.

We visually distinguish these three neuron groups with different colors and positions in the attribution graph view (\autoref{fig:overview}B).
Suppressed neurons are colored \textbf{\blue{blue}} and positioned on the left (\autoref{fig:overview}B.1).
Emphasized neurons are colored \textbf{\orange{orange}} and positioned on the right (\autoref{fig:overview}B.3).
Shared neurons are colored \textbf{\purple{purple}} and positioned in the middle between suppressed neurons and emphasized neurons (\autoref{fig:overview}B.2).
The result is a visualization that disentangles and compares the DNN features and connections from the benign and attacked data.

\paragraph{Fractionation: Characterizing neuron vulnerability.}

Within each group of neurons, we further distinguish them based on their vulnerability.
A neuron is considered more vulnerable if its activation changes greatly under weaker attacks.
We encode neuron vulnerability using its position and color within its group.
More vulnerable neurons are located closer to shared neurons, since they are on the border of the benign and attacked classes, and cause misclassification under weaker attacks. 
Suppressed neurons that are closer to the border with shared neurons are colored purple/blue, and emphasized neurons that are closer to the border with shared neurons are colored purple/orange.

\begin{marginfigure}[2pc]
  \begin{minipage}{\marginparwidth}
    \centering
    \includegraphics[width=1.0\marginparwidth]{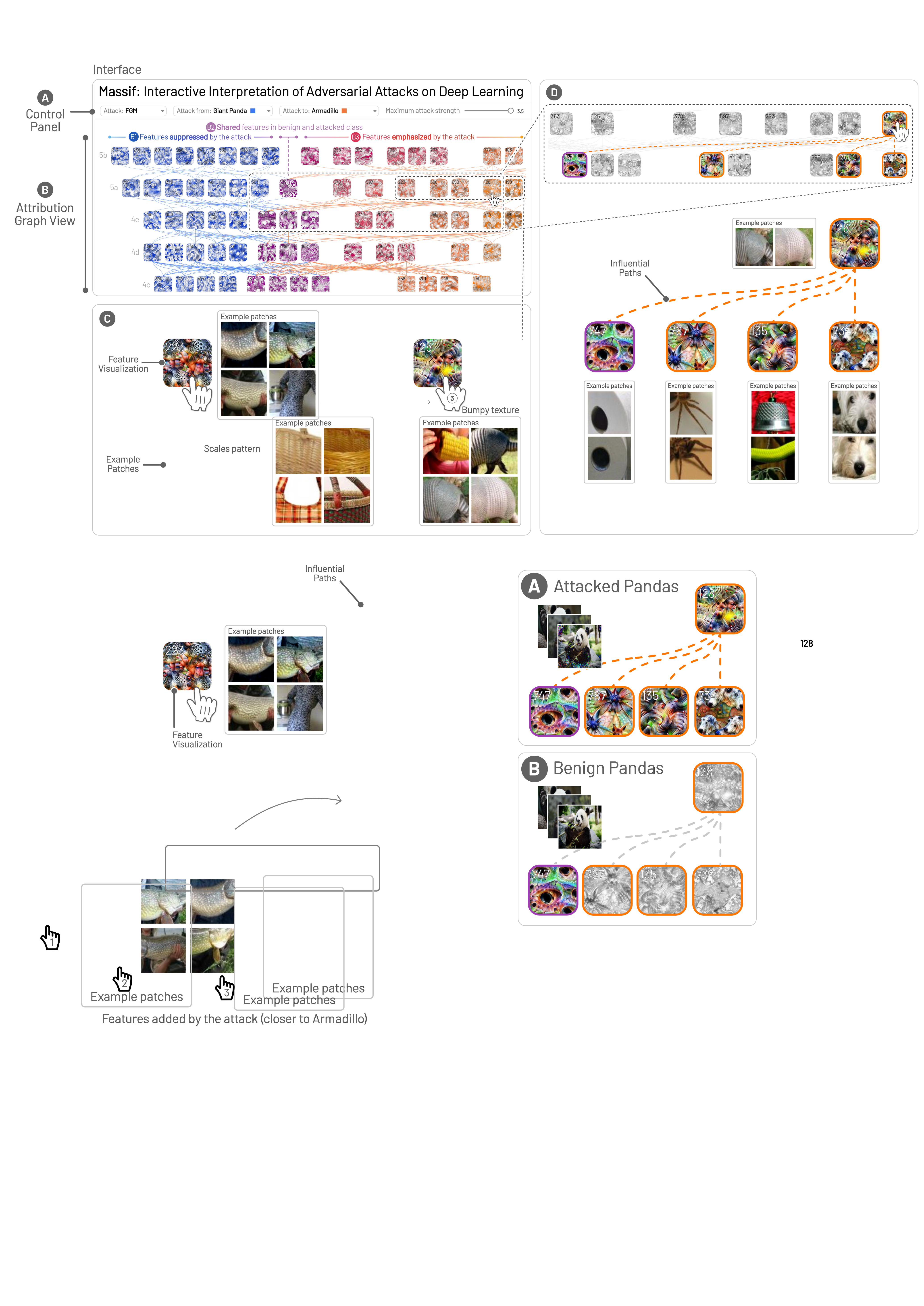}
    \caption{
      Edges in an attribution graph represent influential connections between neurons. \\
      \textbf{(A)} above shows a part of an attribution graph from \autoref{fig:overview}D.
      An emphasized neuron is connected to one shared neuron (\textbf{\purple{purple}}) and three emphasized neurons (\textbf{\orange{orange}}) from the previous layer.
      \textbf{(B)} shows how the same part of the attribution graph looks different for benign images.
      The three emphasized neurons in the previous layer are not activated by the benign inputs, which causes the emphasized neuron in the current layer to be less activated.
    }
    \label{fig:edge}
  \end{minipage}
\end{marginfigure}

\paragraph{Edges: Explaining why features are activated (or suppressed).}
In attribution graphs, edges represent influential connections between neurons that most interact with each other to represent a particular class~\cite{hohman2019summit}.
These connections can explain why an adversarial feature is detected and why some features are suppressed using attribution.

With \model{}, users can interactively visualize attribution, i.e., drilling down into specific subgraphs by hovering over a neuron and highlighting its previous connections which maximally influence the neuron's activation (\autoref{fig:overview}D).
When inspecting a particular emphasized neuron $n$, the highlighted connected neurons from the previous layer will be either shared neurons or other emphasized neurons, as seen in \autoref{fig:edge}A.
Users can observe that the connected emphasized neurons from the previous layer contribute highly to the activation of neuron $n$.
Similarly, to understand why a feature becomes suppressed by an attack, inspecting a particular suppressed neuron $m$ highlights its connected neurons from the previous layer. These will be either shared neurons or suppressed neurons. The neuron $m$ is less activated since the influential neurons from the previous layer are no longer activated, i.e., they are suppressed as well.

%% file: 400-result.tex
\section{Preliminary Results}

We present usage scenarios showing how \model{} can help users better understand adversarial attacks on deep learning models.
Our user Hailey is studying a targeted version of Fast Gradient Method~\cite{Goodfellow2014ExplainingAH} applied on the InceptionV1 model~\cite{szegedy2015going}.
The model is trained on the ImageNet dataset, which contains over 1.2 million images across 1,000 classes.
Using the control panel (\autoref{fig:overview}A), she selects ``giant panda" as the benign class and ``armadillo" as the target class.
She sets the maximum attack strength 3.5.

\paragraph{Which neurons are attacked?} 
Hailey starts by finding which specific neurons are attacked to narrow down the part of the model to investigate.
In the attribution graph view (\autoref{fig:overview}B), she hovers over the suppressed neurons (\autoref{fig:overview}B.1) and the emphasized neurons (\autoref{fig:overview}B.3).
She sees which features are emphasized and suppressed using the neuron feature visualization and dataset example patches.
Exploring these features, she finds the \texttt{mixed5a} layer interesting because three emphasized neurons (223, 698, and 128) in \texttt{mixed5a} look related to armadillo skin patterns (\autoref{fig:overview}C).
She decides to focus on these emphasized neurons in \texttt{mixed5a}.

\paragraph{Which neurons are easily attacked?}
To efficiently devise a countering defense, Hailey wants to prioritize the neurons and investigate them in order.
She knows that \model{} fractionates the neurons according to how easily they are attacked; therefore, she checks how the emphasized neurons in \texttt{mixed5a} are separated (\autoref{fig:overview}C).
By hovering over the neurons from left to right, she observes that ``scales pattern'' is most vulnerable, followed by ``baskets'' and ``bumpy texture'' neurons.
She decides to explore the neurons in this order, since she presumes that it is more efficient to protect more vulnerable neurons.

\paragraph{Why are these neurons attacked?}
Hailey now wants to know how to protect the attacked neurons related to armadillo skin patterns.
She sequentially observes the attribution for the ``scales pattern'', ``baskets'', and ``bumpy texture'' neurons.
Upon inspecting the ``bumpy texture'' neuron (\autoref{fig:overview}D), \model{} shows that four neurons in the previous layer are highly interacting with it: a shared neuron representing ``black circle,'' and emphasized neurons representing ``spider legs'', ``granular texture'', and ``a white hairy dog's face.''
As the three emphasized neurons in the previous layer can be the primary reason behind the detection of ``bumpy texture'', she decides to investigate these neurons more using \model{}.

%% file: 500-conclusion.tex
\section{Ongoing Work}

\paragraph{Interactive neuron editing.}
\model{} currently visualizes the neurons that are activated or suppressed by an attack under varying degrees of severity.
We are working on extending \model{}'s interactivity by allowing real-time neuron editing, e.g., deletion.
This would allow a user to actively identify vulnerable neurons using our visualization and interactively remove them from the DNN to observe its effect in real-time.
Neuron deletion would mask the activations of a particular neuron, potentially preventing the malicious effect of a targeted attack to propagate deeper into the network.
This would enable a user to preemptively edit a DNN to enhance its robustness to adversarial attacks.
For example, a user may identify and choose to delete a shared neuron that only feeds into emphasized neurons, preventing adversarially activated neurons from having any effect in the subsequent layers of the network, thus thwarting the targeted attack from succeeding.

\paragraph{Planned evaluation.}
We plan to evaluate the effectiveness of our visualization tool coupled with interactive neuron editing through in-lab user studies where participants seek to increase the robustness of a large-scale, pretrained DNN model.
We will recruit students with basic knowledge of deep learning models.
All participants will be asked edit the DNN for different benign-attacked class pairs and will be evaluated on the basis of reduction in targeted attack success rate.
We will also conduct pre-test and post-test surveys to evaluate whether \model{} gave any deeper insights into the failure modes of the studied DNN and what factors the participants considered while editing the DNN to increase its robustness to adversarial attacks.

\section{Conclusion}
We present \model{}, an interactive system we are developing that visualizes how adversarial attacks permeate DNN models and cause misclassification.
\model{} generates and visualizes multiple attribution graphs as a summary of what features are important for a particular class (e.g., benign or attacked class) and how the features are related.
\model{} enables flexible comparison between benign and attacked attribution graphs, highlighting where and why the attribution graphs start to diverge, ultimately helping people better interpret complex deep learning models, their vulnerabilities, and how to best construct defenses.

%% file: main.bbl

\begin{thebibliography}{00}

%%% the Web address.

\ifx \showCODEN    \undefined \def \showCODEN     #1{\unskip}     \fi
\ifx \showDOI      \undefined \def \showDOI       #1{{\tt DOI:}\penalty0{#1}\ }
  \fi
\ifx \showISBNx    \undefined \def \showISBNx     #1{\unskip}     \fi
\ifx \showISBNxiii \undefined \def \showISBNxiii  #1{\unskip}     \fi
\ifx \showISSN     \undefined \def \showISSN      #1{\unskip}     \fi
\ifx \showLCCN     \undefined \def \showLCCN      #1{\unskip}     \fi
\ifx \shownote     \undefined \def \shownote      #1{#1}          \fi
\ifx \showarticletitle \undefined \def \showarticletitle #1{#1}   \fi
\ifx \showURL      \undefined \def \showURL       #1{#1}          \fi

\bibitem{carter2019activation}
{Shan Carter}, {Zan Armstrong}, {Ludwig Schubert}, {Ian Johnson}, {and} {Chris
  Olah}. 2019.
\newblock \showarticletitle{Activation atlas}.
\newblock {\em Distill\/} {4}, 3 (2019), e15.
\newblock


\bibitem{chen2018shapeshifter}
{Shang-Tse Chen}, {Cory Cornelius}, {Jason Martin}, {and} {Duen Horng~Polo
  Chau}. 2018.
\newblock \showarticletitle{Shapeshifter: Robust physical adversarial attack on
  faster r-cnn object detector}. In {\em Joint European Conference on Machine
  Learning and Knowledge Discovery in Databases}. Springer, 52--68.
\newblock


\bibitem{esteva2019guide}
{Andre Esteva}, {Alexandre Robicquet}, {Bharath Ramsundar}, {Volodymyr
  Kuleshov}, {Mark DePristo}, {Katherine Chou}, {Claire Cui}, {Greg Corrado},
  {Sebastian Thrun}, {and} {Jeff Dean}. 2019.
\newblock \showarticletitle{A guide to deep learning in healthcare}.
\newblock {\em Nature medicine\/} {25}, 1 (2019), 24.
\newblock


\bibitem{Goodfellow2014ExplainingAH}
{Ian~J. Goodfellow}, {Jonathon Shlens}, {and} {Christian Szegedy}. 2014.
\newblock \showarticletitle{Explaining and Harnessing Adversarial Examples}.
\newblock {\em CoRR\/}  {abs/1412.6572} (2014).
\newblock


\bibitem{grigorescu2019survey}
{Sorin Grigorescu}, {Bogdan Trasnea}, {Tiberiu Cocias}, {and} {Gigel Macesanu}.
  2019.
\newblock \showarticletitle{A survey of deep learning techniques for autonomous
  driving}.
\newblock {\em Journal of Field Robotics\/} (2019).
\newblock


\bibitem{guo2019survey}
{Guodong Guo} {and} {Na Zhang}. 2019.
\newblock \showarticletitle{A survey on deep learning based face recognition}.
\newblock {\em Computer Vision and Image Understanding\/}  {189} (2019),
  102805.
\newblock


\bibitem{hohman2019summit}
{Fred Hohman}, {Haekyu Park}, {Caleb Robinson}, {and} {Duen~Horng Chau}. 2019.
\newblock \showarticletitle{Summit: Scaling Deep Learning Interpretability by
  Visualizing Activation and Attribution Summarizations}.
\newblock {\em IEEE VIS\/} (2019).
\newblock


\bibitem{kurakin2016adversarial}
{Alexey Kurakin}, {Ian Goodfellow}, {and} {Samy Bengio}. 2016.
\newblock \showarticletitle{Adversarial examples in the physical world}.
\newblock {\em arXiv preprint arXiv:1607.02533\/} (2016).
\newblock


\bibitem{nassif2019speech}
{Ali~Bou Nassif}, {Ismail Shahin}, {Imtinan Attili}, {Mohammad Azzeh}, {and}
  {Khaled Shaalan}. 2019.
\newblock \showarticletitle{Speech recognition using deep neural networks: A
  systematic review}.
\newblock {\em IEEE Access\/}  {7} (2019), 19143--19165.
\newblock


\bibitem{olah2017feature}
{Chris Olah}, {Alexander Mordvintsev}, {and} {Ludwig Schubert}. 2017.
\newblock \showarticletitle{Feature Visualization}.
\newblock {\em Distill\/} (2017).
\newblock
\showDOI{%
\url{http://dx.doi.org/10.23915/distill.00007}}
\newblock
\shownote{https://distill.pub/2017/feature-visualization.}


\bibitem{qin2019imperceptible}
{Yao Qin}, {Nicholas Carlini}, {Ian Goodfellow}, {Garrison Cottrell}, {and}
  {Colin Raffel}. 2019.
\newblock \showarticletitle{Imperceptible, robust, and targeted adversarial
  examples for automatic speech recognition}.
\newblock {\em arXiv preprint arXiv:1903.10346\/} (2019).
\newblock


\bibitem{ross2018improving}
{Andrew~Slavin Ross} {and} {Finale Doshi-Velez}. 2018.
\newblock \showarticletitle{Improving the adversarial robustness and
  interpretability of deep neural networks by regularizing their input
  gradients}. In {\em Thirty-second AAAI conference on artificial
  intelligence}.
\newblock


\bibitem{szegedy2015going}
{Christian Szegedy}, {Wei Liu}, {Yangqing Jia}, {Pierre Sermanet}, {Scott
  Reed}, {Dragomir Anguelov}, {Dumitru Erhan}, {Vincent Vanhoucke}, {and}
  {Andrew Rabinovich}. 2015.
\newblock \showarticletitle{Going deeper with convolutions}. In {\em
  Proceedings of the IEEE conference on computer vision and pattern
  recognition}. 1--9.
\newblock


\bibitem{tao2018attacks}
{Guanhong Tao}, {Shiqing Ma}, {Yingqi Liu}, {and} {Xiangyu Zhang}. 2018.
\newblock \showarticletitle{Attacks meet interpretability: Attribute-steered
  detection of adversarial samples}. In {\em Advances in Neural Information
  Processing Systems}. 7717--7728.
\newblock


\end{thebibliography}
